\documentclass[10pt,twocolumn,letterpaper]{article}

\usepackage{cvpr}              

\usepackage[nolist]{acronym}
\usepackage{adjustbox}
\usepackage{amsmath}
\usepackage{amssymb}
\usepackage{amsthm}
\usepackage{array}  
\usepackage{booktabs}
\usepackage{capt-of}
\usepackage[font={small}]{caption}
\usepackage{subcaption}
\usepackage{floatrow}
\usepackage{colortbl}
\usepackage{wrapfig}
\usepackage{graphicx}
\usepackage{multirow}
\usepackage{paralist}
\usepackage{tabularx}
\usepackage[table, xcdraw]{xcolor}

\definecolor{mathpurple}{rgb}{0.73, .58, .92}
\definecolor{stringsgreen}{rgb}{.5, .72, 0.42}
\definecolor{entityyellow}{rgb}{1 .83, .15}
\definecolor{relationblue}{rgb}{.42, .64, .82}
\definecolor{highlight}{rgb}{0.824,0.976,0.824}

\definecolor{flodarkpurple}{rgb}{0.288,0.1196,0.7}

\definecolor{amber}{rgb}{1.0, 0.75, 0.0}

\newcommand{\methodnamefull}{Semantic Graph Consistency\xspace}
\newcommand{\methodname}{SGC\xspace}

\definecolor{cvprblue}{rgb}{0.21,0.49,0.74}
\usepackage[pagebackref,breaklinks,colorlinks,citecolor=cvprblue]{hyperref}

\def\paperID{9084} 
\def\confName{CVPR}
\def\confYear{2024}

\title{Semantic Graph Consistency: Going Beyond Patches for Regularizing Self-Supervised Vision Transformers}

\author{Chaitanya Devaguptapu$^*$ \hspace{0.5cm} Sumukh K Aithal$^*$ \hspace{0.5cm} Shrinivas Ramasubramanian\\
Yamada Moyuru \hspace{1cm} Manohar Kaul \\\\
\textsuperscript{}Fujitsu Research India\\
}

\begin{document}

\begin{acronym}
\end{acronym}

\maketitle
\def\thefootnote{*}\footnotetext{These authors contributed equally to this work. For questions, reach out to email@chaitanya.one}

\begin{abstract}
Self-supervised learning (SSL)  with vision transformers (ViTs) has proven effective for representation learning as demonstrated by the impressive performance on various downstream tasks. Despite these successes, existing ViT-based SSL architectures do not fully exploit the ViT backbone, particularly the patch tokens of the ViT. In this paper, we introduce a novel Semantic Graph Consistency (SGC) module to regularize ViT-based SSL methods and leverage patch tokens effectively. 
We reconceptualize images as graphs, with image patches as nodes and infuse relational inductive biases by explicit message passing using Graph Neural Networks into the SSL framework. Our SGC loss acts as a regularizer, leveraging the underexploited patch tokens of ViTs to construct a graph and enforcing consistency between graph features across multiple views of an image. Extensive experiments on various datasets including ImageNet, RESISC and Food-101 show that our approach significantly improves the quality of learned representations, resulting in a 5-10\% increase in performance when limited labeled data is used for linear evaluation. These experiments coupled with a comprehensive set of ablations demonstrate the promise of our approach in various settings.


\end{abstract}

\vspace{-0.3cm}
\section{Introduction}
The increasing need for extensive labeled data in training deep learning models has spurred interest in self-supervised learning (SSL) methods. Self-supervised learning, distinct from traditional supervised learning, leverages the inherent structure of data to generate surrogate labels, eliminating the need for explicit annotations. Predominant SSL techniques learn by minimizing errors in pretext tasks. These tasks, such as predicting the sequence of image patches~\cite{doersch2015unsupervised} or the rotation of an image~\cite{GidarisSK18}, are deliberately challenging to encourage the extraction of meaningful data features. However, designing effective pretext tasks is non-systematic and demands substantial engineering. On the other hand, contrastive learning a more recent and a prominent method for SSL operates by learning representations such that similar or ``positive'' pairs of data points are brought closer together in the representation space, while dissimilar or ``negative'' pairs are pushed further apart. This framework encompasses methods methods like SimCLR \cite{chen2020simple} and MoCo \cite{he2020momentum}. \\
\begin{figure*}
    \centering
    \includegraphics[width=0.7\linewidth]{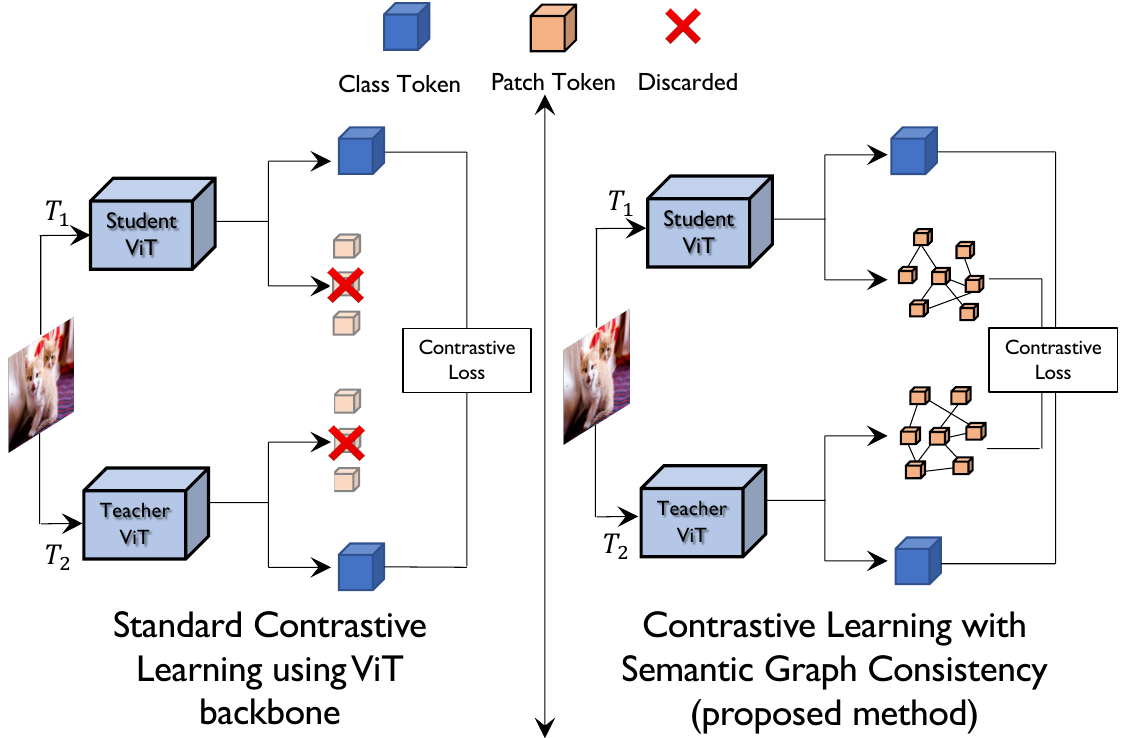}
    \caption{Conceptual representation of the proposed \methodnamefull (\methodname). \textit{\textbf{Unlike traditional methods that emphasize on class token representations and discard patch tokens for contrastive learning, \textit{\textbf{\methodname}} leverages patch tokens to accentuate relational and semantic information}}. SGC constructs a graph utilizing the patch tokens and imposes graph-level consistency, thus significantly enhancing the representational quality within the contrastive learning framework.}
    \label{fig:intro-figure}
    \captionsetup{belowskip=-10pt}

\end{figure*}

\noindent While state-of-the-art contrastive learning methods for SSL adopt Vision Transformers (ViTs) as their backbone, their utilization of the rich representations learned by ViTs remains sub-optimal. ViTs transform the images into a sequence of tokens and performs a set of attention operations sequentially. The final output of a ViT includes a class token ([CLS]) for global representation and patch tokens corresponding to each image patch that is passed as input. Predominantly, ViT-based SSL methods focus on the class token, overlooking the informative content in patch tokens. This oversight is significant, as patch tokens encapsulate essential details not present in the class token, as demonstrated in \cite{naseer2021intriguing}. \citet{jiang2021all} address this by introducing `token labeling' during the training phase of ViTs, employing both patch and class tokens to capture rich local information present in patch tokens effectively. Notably, token labeling~\cite{jiang2021all} is primarily applied in a supervised learning context, necessitating labels for training. In contrast, our work focuses on the SSL setting where such explicit labels are not available and thus, presents a different set of challenges and opportunities. 


\noindent Our approach reconceptualizes images as graphs, capturing intricate relationships between image patches in a manner that is reflective of how the human visual system processes and interprets visual scenes. This perspective draws inspiration from the human visual system's remarkable ability to discern semantic relationships between objects in a scene almost instantaneously~\cite{wolfe1998visual}. In the proposed approach, every patch is treated as a node in a semantic graph that is constructed using the patch tokens.
Calls for image representation based on part-whole hierarchies~\cite{hinton2022represent, han2022vision, ding2023visual} further underscore the potential of graph-based representations, emphasizing the need for models that can leverage the hierarchical and relational structure inherent in visual data. 

\noindent Building on these insights, we introduce a novel approach \methodnamefull ({\methodname}) to improve the representations learned by existing ViT based SSL approaches. The proposed approach helps learn better representation by serving as a regularizer, fostering the learning of discriminative and semantically meaningful features.  \methodnamefull enforces consistency between the graphs constructed from the different views of an image. We use Graph Neural Networks (GNNs) to perform message passing on this constructed graph and compute the \textit{graph consistency} loss. We observe that the addition of SGC module leads to improved performance compared to the base SSL method on various datasets. This is because the patch tokens (used to construct the graph) are often discarded by the existing ViT-SSL approaches but contains localized and fine-grained knowledge about the image which can guide the SSL model towards capturing the right set of features. 
Moreover, treating images as graphs allows SSL models to capture hierarchical relationships between patches, mirroring the inherent structure in visual data. 
We hypothesize that this guides the model towards learning more generalizable and robust features akin to how humans effortlessly understand the context and relationships between objects in a scene. The core-idea of our approach along with how it differs from existing approaches is shown in Fig \ref{fig:intro-figure}. We summarize our contributions below:

\begin{compactitem}
    \item Our work primarily focuses on an regularizing SSL with the goal of making effective use of the patch tokens in Vision Transformers.
    \item We introduce a novel {\methodnamefull}({\methodname}) module, that leverages the patch tokens with GNNs to infuse relational inductive biases into the SSL framework.
    \item The addition of {\methodname} regularizer significantly improves the performance of the base SSL method on six datasets, in two different settings, with massive gains in the linear evaluation with limited data setup. In fact, {\methodname} results in 5-10 \% increase in accuracy when 1-5\%  data is used for pre-training.
\end{compactitem}
\vspace{-0.3cm}
\section{Related Work}
\vspace{-0.3cm}

\paragraph{Transformers for Vision tasks:} Vision Transformers~\cite{dosovitskiy2021an,pmlr-v139-touvron21a,liu2021Swin}, are based on the transformer architecture \cite{vaswani2017attention} which was originally developed for natural language processing tasks. The core idea of ViT is to treat an image as a sequence of patches, which are then fed into a Transformer encoder to extract high-level features that are used for various tasks like image classification, object detection etc. ViTs achieve state-of-the-art performance on complex and large scale datasets like ImageNet~\cite{imagenet}. Various works~\cite{wang2021pyramid, NEURIPS2021_4e0928de, dai2021coatnet, wu2021cvt, huang2021shuffle,DBLP:conf/iclr/0001YCL00L22} have been proposed to make ViTs more efficient in terms of compute, data requirements and inference speeds. 
\paragraph{Self-Supervised Learning using Vision Transformers:} Self-supervised learning (SSL) focuses on learning to perform a task without explicitly labelling the data for that task~\cite{doersch2015unsupervised,agrawal2015learning,wang2015unsupervised,caron2020unsupervised}.
Given an unalabelled dataset, SSL models learn to predict how to fill in missing pixels in an image or aims to learn invariant semantics of two random views of an image~\cite{tian2019contrastive}. By performing these tasks (often referred to as pretext tasks), the model learns a feature representation that helps in solving various downstream tasks, such as classification~\cite{azizi2021big} or object detection~\cite{dang2022study}. SSL helps in learning the latent representation of the data without incurring the cost of manually annotating the data samples.  SSL has been shown to be effective in learning representations of data that are more robust to noise and variations in the data distribution \cite{hendrycks2019using}. This makes self-supervised learning a promising approach for training models that can generalize well to new data.
\\
\noindent Vision Transformers have been proven to be a powerful and versatile architecture that is well-suited for SSL~\cite{bao2021beit,he2021masked,xie2021simmim,caron2021emerging}. ViTs help in learning long range dependencies in images, which is important for SSL tasks that require understanding of the global context of an image. Most of the SSL works design the pretext task in an architecture agnostic manner. There have very less efforts on leveraging the architectural advantages of ViTs for improving Self supervision. Methods like Self-Patch~\cite{yun2022patch} and iBOT~\cite{zhouimage} make an attempt to exploit patch token in the SSL setting, enhancing downstream task performance. However, Self-Patch limits its focus to neighboring patch interactions, missing global patch dynamics. Conversely, iBOT averages loss between masked and unmasked views of the patch tokens, losing on the information of the relationship between patches. As a result, there is a potential opportunity to enhance their performance further by effectively incorporating the full set of learned representations into the SSL framework, unlocking the true potential of vision transformers in SSL for visual recognition tasks. In contrary to these existing approaches, we take a different route and our methods differ significantly. We build a graph and leverage a GCN~\cite{kipf2017semisupervised} to learn the relationship between the patches explicitly.


\vspace{-0.5cm}
\paragraph{Graphs in Computer Vision: }
Graphs have been used in classical computer vision algorithms like Normalized Cut~\cite{shi2000normalized} which poses image segmentation as a graph partitioning problem. An image can be represented as a spatial graph where the pixels (or superpixels~\cite{wang2011superpixel}) are the nodes in the graph. Similarly, a video can be represented as a spatio-temporal graph~\cite{wang2018videos}. Scene graphs \cite{liang2016semantic} parse the image into a semantic graph which captures the relationship between objects. Graph neural networks (GNNs) \cite{zhou2020graph, wu2020comprehensive,kipf2016semi,kipf2017semisupervised} learn a feature representation for every node in the graph with the help of neural networks. The node representation are usually updated with the aggregated information from the node's neighbours in message passing type GNNs. More recently, these GNNs have been extensively used in specific computer vision tasks like semantic segmentation \cite{li2020spatial}, action recognition \cite{yan2018spatial, chen2019graph} and visual question answering \cite{li2019relation, narasimhan2018out, teney2017graph} to name a few. Vision-GNN~\cite{han2022vision} is a GNN based backbone for computer vision has been shown to achieve good performance on visual recognition benchmarks. Graphs induce a relational inductive bias into the model and this property makes it an appealing choice for various vision tasks. To the best of our knowledge, we are the first to introduce this relational inductive bias of Graphs in Self-supervised visual representation learning.
\begin{figure*}
    \centering
\includegraphics[width=0.9\textwidth]{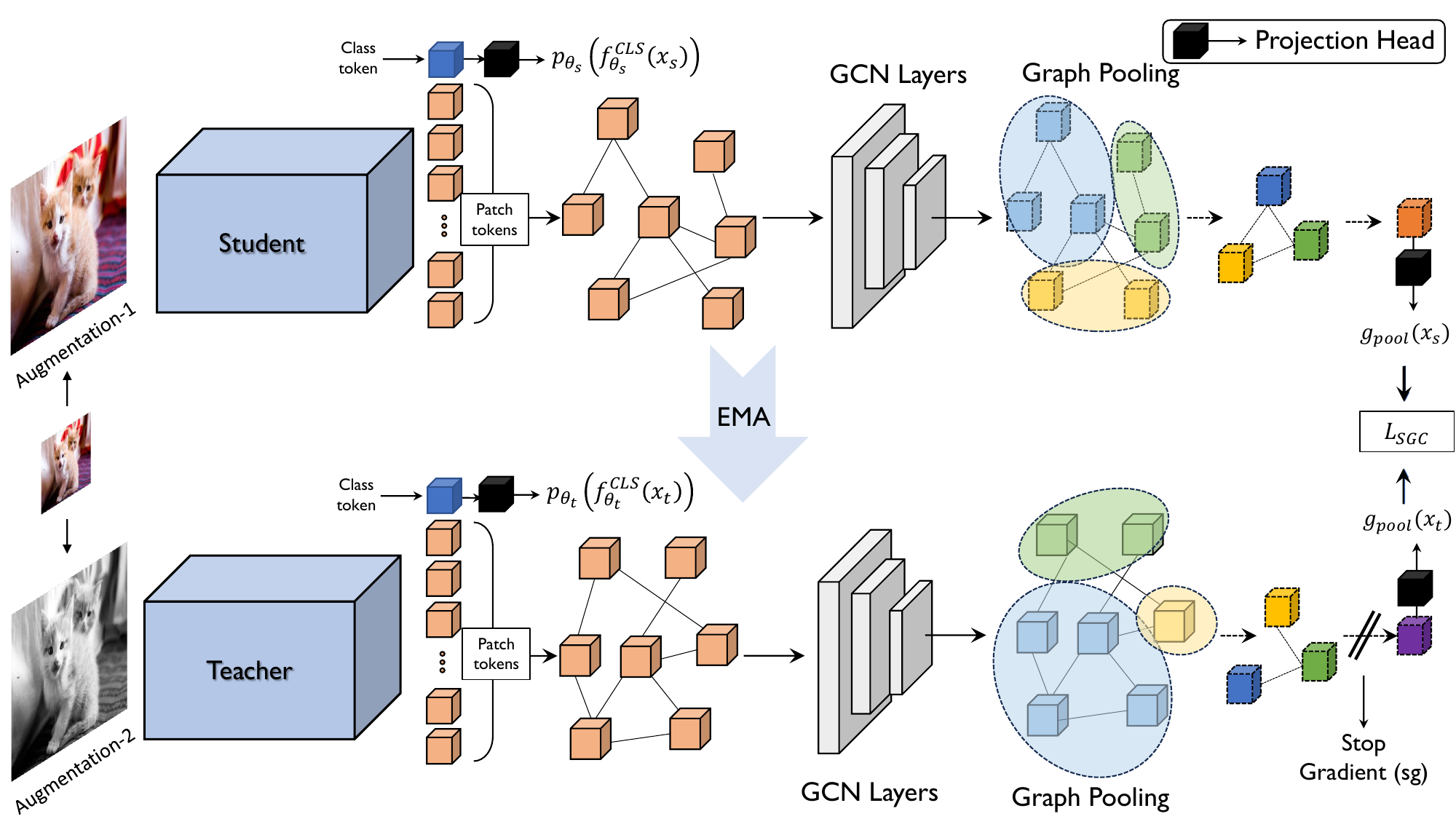}
    \caption{Overview of the proposed \methodnamefull for Contrastive learning using Vision Transformer backbone. (EMA denotes exponential moving average.)} 
    \label{fig:framework}
\end{figure*}

\vspace{-0.3cm}
\section{Method}
\vspace{-0.3cm}

In this section, we delve into the specifics of our methodology for inducing explicit semantic structure in self-supervised ViTs. We commence with a brief overview of ViTs and a summary of ViT-based SSL methods, as they form the bedrock of our research. Subsequently, we elucidate our proposed approach, emphasizing the steps undertaken to introduce graph consistency for regularizing ViT based SSL approaches.

\subsection{Preliminaries}
\paragraph{Vision Transformer:} Consider an image $ \mathbf{x} \in \mathbb{R}^{H \times W \times C}$, where $(H, W)$ represent the height and width of the image, respectively, and $C$ denotes the number of channels in the image (typically, $C=3$ for RGB images). ViTs \cite{dosovitskiy2021an, vit-survey} treat images as a sequence of patches and partition the image into $N$ non-overlapping square patches $\mathbf{x}_{{i \dots N}} \in \mathbb{R}^{P \times P}$, where $P$ signifies the patch size. Each patch is transformed into a feature vector $\mathbf{v}_{{i \dots N}} \in \mathbb{R}^{D}$ via a linear projection, and a positional embedding is added (based on the relative position of the patch in the image). Here, $D$ denotes the feature dimension of the final embedding. An embedding of the class token, $v_{\text{[CLS]}}$, is appended to provide a global view of the image. The final input sequence passed into the ViTs is denoted as $ [v_{\text{[CLS]}}, v_{1}, v_{2}, \dots, v_{N}]$. 
\begin{align}
f_{\theta}(\mathbf{x}) = f_{\theta}( [v_{\text{[CLS]}}, v_{1}, v_{2}, \dots, v_{N}])
\end{align}
ViT learns a representation for each of the input tokens. $f_{\theta}(v_{\text{[CLS]}})$ denotes the representation learned for the class token, and $f_{\theta}(v_{i})$ denotes the representation learned for the $i$-th patch token. $f_{\theta}(v_{\text{[CLS]}})$ is typically used for tasks that require a global understanding of the image, while $f_{\theta}(v_{i})$ is used for tasks that necessitate pixel or patch-level understanding.
\vspace{-0.5cm}

\paragraph{Self-supervised Vision Transformers:} ViTs have emerged as the preferred choice for many recent SSL methodologies \cite{caron2021emerging,yun2022patch,chen2021empirical,xie2021selfsupervised}. Some of the most successful approaches, such as DINO~\cite{caron2021emerging}, leverage contrastive learning for SSL on ViTs. By applying different augmentations to the same image, a pair of representations are constructed. An objective function ($\mathcal{L_{\text{SSL}}}$) is then designed to learn an invariant representation between the two augmentations.
\\
\noindent Let $\mathbf{x}_s$ and $\mathbf{x}_t$ denote the two augmented versions of an input image $\mathbf{x}$ passed to the student and teacher networks of a ViT-based SSL framework, respectively. SSL approaches based on contrastive learning aim to learn an invariant representation by minimizing the distance (denoted by $H$) between the projected representations learned by the class token of both views. The objective of the ViT-based SSL approaches can be calculated as follows:
\begin{equation}
\mathcal{L_{\text{SSL}}} =  H\Big[p_{\theta_{s}}\big(f_{\theta_s}^{\text{CLS}}(\mathbf{x}_s)\big), \texttt{sg}\big[p_{\theta_{t}}\big(f_{\theta_t}^{\text{CLS}}(\mathbf{x}_t)\big)\big]\Big]
\label{eq:cls_eq}
\end{equation}
\noindent Here, $\theta_s$, $\theta_t$ denote the parameters of the student and the teacher network respectively. $\texttt{sg}$ denotes the stop-gradient operation and $p$ denotes a projection head on top of the class token representations. Parameters of the teacher network $\theta_t$ are updated using exponential moving average of $\theta_s$. The choice of the distance function depends on the type of self-supervision, with the Kullback-Leibler divergence being the most common choice for $H$. Notably, most ViT-based SSL approaches rely solely on the class token of the ViT for contrastive learning and do not leverage the patch tokens. Our primary contribution is the effective utilization of these patch tokens to regularize the representation learning using ViT based SSL approaches.
\vspace{-0.2cm}

\subsection{\methodnamefull}

\paragraph{Graph Construction:} ViTs learn a representation for each of the patch tokens. For simplicity, we use $v_i$ to denote the representation $f_{\theta}(v_i)$, learned for the $i$-th patch token.
The output of the ViT comprises a class token and $N$ patch tokens. These patch tokens can be viewed as a set of unordered nodes, denoted as $\mathcal{V}={v_1,v_2,\cdots,v_N}$. We construct a k-nearest neighbors (KNN) graph using the representations learned by the patch tokens, where each patch token has $k$ neighbors. For each node $v_i$, we identify its $k$ nearest neighbors, $\mathcal{N}(v_i)$ and add an edge $e_{ij}$ from $v_i$ to $v_j$ for all $v_j \in \mathcal{N}(v_i)$. This results in a graph $\mathcal{G}=(\mathcal{V},\mathcal{E})$, where $\mathcal{E}$ represents all the edges. In our work, we use the Euclidiean distance and/or Cosine Similarity to measure the similarity of feature representations while constructing the Nearest Neighbor graph. This explicit construction of the graph allows us to perceive the image as a graph. Our choice of using nearest neighbors for constructing the graph is primarily based on Vision-GNN~\cite{han2022vision}.

\paragraph{Message Passing using GNNs:} Graph Neural networks (GNNs) exploit the inherent structure of graph data and capture the dependencies between connected nodes. GNNs can be understood in terms of two fundamental operations: aggregation and update.  Consider a graph $G = (V, E)$, where $V$ denotes the set of nodes and $E$ denotes the set of edges. Each node $v_i$ has a feature vector $x_i$.  The aggregation operation in GNNs is responsible for collecting information from a node's local neighborhood. For a given node $v_i$, the aggregation operation collects the feature vectors of its neighboring nodes and combines them. This is typically done by summing the feature vectors, but other aggregation functions such as mean, max, or even more complex functions can be used. Mathematically, this can be represented as:
\begin{equation}
    h_{N(i)}^{(l)} = \text{AGGREGATE}^{(l)}(\{h_{j}^{(l)}: j \in N(i)\})
\end{equation}

\noindent where, $h_{j}^{(l)}$ is the feature vector of node $j$ at layer $l$, $N(i)$ is the set of neighbors of node $i$, and $\text{AGGREGATE}^{(l)}$ is the aggregation function at layer $l$.
The update operation takes the aggregated information from a node's neighborhood and uses it to update the node's own feature vector. This is typically done by transforming the aggregated vector using a learned weight matrix and applying a non-linear activation function. Mathematically, this can be represented as:
\begin{equation}
    h_{i}^{l+1} = \sigma(W^{(l)} h_{N(i)}^{(l)})
\end{equation}
where $W^{l}$ is the weight matrix at layer 
$l$, and $\sigma$ is a non-linear activation function. These two operations, repeated over multiple layers, allow GNNs to propagate and transform information across the graph, capturing both local and global structural patterns in the graph. Various GNN layers such as GCN \cite{kipf2016semi}, GIN \cite{xu2018powerful} and SAGE \cite{hamilton2017inductive} can be defined based on the update and aggregate operations. We analyse the impact of GNN layers in the ablation study (Table \ref{tab:gnn_layer_ablation}). The KNN graph constructed from the patch tokens is then passed through a set of graph neural network layers to compare the two view of the image via explicit semantic understanding. 
\vspace{-0.5cm}
\paragraph{Connection to Weisfeiler-Lehman algorithm}: We now answer the question of how the graph neural networks is used to compute the graph consistency.
The Weisfeiler-Lehman (WL) algorithm \cite{weisfeiler1968reduction} is a classic technique for graph isomorphism testing that operates through color refinement. It has been shown that a few layers of GCN is equally powerful as the 1-WL graph isomorphism test \cite{xu2018powerful, chen2019equivalence}. Interestingly, our \methodnamefull  draws a conceptual parallel to the WL algorithm in terms of graph consistency.  By ensuring that the constructed graph representations across different views of an image are similar, we effectively aim to establish structural coherence between the two representations with GNNs, akin to the WL algorithm's role in recognizing isomorphism between graphs. This conceptual connection highlights the significance of our graph consistency approach and underscores its potential in capturing meaningful structural information in SSL settings.

\vspace{-0.5cm}

\paragraph{Pooling: }Graph Pooling ~\cite{ying2018hierarchical, bianchi2019mincut} reduces the graph size while retaining significant structural and feature information. Initially, each node is assigned a score, typically computed as a function of the node's features, such as $s_i = \sigma(v_{i}^{T} W)$, where $W$ is a learnable weight vector, $v_i$, is the node's feature vector and $\sigma$ is the sigmoid function. The top-$k$ nodes based on these scores are selected to form a coarsened graph. The adjacency and feature matrices of this coarsened graph are updated to reflect the new set of nodes and their connections. 
Pooling allows operating at different scales and helps capture hierarchical patterns in the graph. This is analogous to capturing relationship between parts of an object in the first layer, interaction between different objects in the second layer and object-scene interaction in the final layer.
\vspace{-0.5cm}

\paragraph{{\methodname} as a Regularizer:} Our objective function build on the top of the learned patch tokens, message passing and graph pooling. We now define our objective function $L_{\text{SGC}}$. For a given input $\mathbf{x}$, the graph pooled features can be calculated as shown below. 
\begin{align}
    \label{eq:knn} G(V, E)(\mathbf{x}) &= \text{KNN}\Big[f_{\theta}^{[v_{1}, v_{2}, \dots, v_{N}]}(\mathbf{x})\Big] \\
    \label{eq:g_conv} g_{f}(\mathbf{x}) &= \text{GraphConv}(G(V, E)(\mathbf{x})) \\ 
    \label{eq:g_pool} g_{\text{pool}}(\mathbf{x}) &= \text{Pool}(g_{f}(\mathbf{x}))
\end{align}
We first compute the patch token representations for a given input and build a KNN-graph using the patch tokens (Eq. \ref{eq:knn}), then we use a set of GNN layers (denoted by GraphConv) to do message passing on top of the constructed KNN-graph (Eq. \ref{eq:g_conv}) and pass the learned features through the graph pooling layers (denoted by Pool in Eq. \ref{eq:g_pool}). We obtain the graph pooled representations for both the views passed to the student and the teacher and pass them to the projection head (denoted using $g_f$), the final {\methodname} loss is computed as shown below: 
\begin{equation}
    \mathcal{L_{\text{\methodname}}} =  H\Big[p_{\theta_{s}}\big(g_{\text{pool}}(\mathbf{x}_s)\big), \text{sg}\big[p_{\theta_{t}}\big(g_\text{pool}(\mathbf{x}_t)\big)\big]\Big]
    \label{eq:sgc-eq}
\end{equation}
Our end-to-end pipeline is also summarised in Figure-\ref{fig:framework}. 
\vspace{-0.5cm}

\paragraph{Overall Objective Function:} The primary objective of most existing ViT-based SSL methods is to minimize the distance between the representations learned by the class tokens of the student and teacher networks. This is achieved by defining a distance metric on top of the ViT's class tokens and optimizing the weights with the objective of reducing this distance, as described in Equation \ref{eq:cls_eq}. In our proposed {{\methodname}} method, we extend this objective by introducing an additional term that aims to minimize the distance between the pooled graph features. This additional term acts as a regularizer, ensuring that the learned representations are more robust and semantically meaningful. Mathematically, our overall objective function can be represented as:

\begin{equation}
\mathcal{L} =  \mathcal{L}_{\text{SSL}} + \beta \times \mathcal{L}_{\text{SGC}}
\label{eq:overall-objective}
\end{equation}

\noindent Here, \(\mathcal{L}_{\text{SSL}}\) denotes the loss associated with the class tokens, which is commonly used in existing ViT-based SSL methods. \(\mathcal{L}_{\text{\methodname}}\) represents the loss associated with the pooled graph features, introduced in our method. The coefficient \(\beta\) is a hyperparameter that control the relative importance of the \methodname\xspace loss in the overall objective. By optimizing this combined objective, our method not only learns representations that are invariant to augmentations (as in traditional SSL methods) but also ensures that these representations capture the structural semantics of the image, leading to improved performance on small-scale datasets. In essence, while the first term \(\mathcal{L}_{\text{SSL}}\) ensures global consistency between the augmented views of an image, the second term \(\mathcal{L}_{\text{\methodname}}\) enforces local consistency by leveraging the structural information captured in the graph. This dual-objective approach ensures that our method benefits from both global and local semantic cues, leading to richer and more discriminative representations.

\begin{table*}[h]
\begin{tabular}{>{\kern-\tabcolsep}l|cccccc<{\kern-\tabcolsep}}
\toprule
             & ImageNet-25    & RESISC         & Caltech-256    & Food-101             & \multicolumn{1}{l}{ImageNet-100} & ImageNet-1K          \\ \midrule
Pre-training Epochs             & 300    & 300         & 300    & 300             & 300 & 100          \\ \midrule

BYOL~\cite{grill2020bootstrap}         & 74.60          & 82.70           & 38.51          & 66.49                & 71.58                            & 59.44                \\
SimCLR~\cite{chen2020simple}       & 69.01          & 71.35          & 33.56          & 64.15                & \textbf{74.26}                            & 56.77                \\
Barlow Twins~\cite{zbontar2021barlow} & 72.76          & 77.62          & 40.86          & 62.60                 & 68.18                            & 62.62                \\
DINO~\cite{caron2021emerging}         & 77.08          & 87.76          & 45.31          & 65.08       & 66.40                    & 63.49       \\ \midrule
\rowcolor{highlight} DINO+\methodname     & \textbf{79.47} & \textbf{88.32} & \textbf{46.82} & \textbf{67.97} & {71.90}             & \textbf{64.32} \\ \bottomrule
\end{tabular}
\caption{\textit{\textbf{Augmenting SoTA approach like DINO with \methodname improves it's performance}}. Performance comparison of the proposed method \methodnamefull (\methodname) against leading self-supervised learning (SSL) techniques with a Vision Transformer (ViT) backbone using the Lightly benchmark\cite{susmelj2020lightly}}
\label{tab:lightly-benchmark}
\end{table*}

\vspace{-0.3cm}

\section{Experiments and Results}

In this section, we conduct a systematic evaluation of our proposed \methodnamefull (\methodname), shedding light on its performance across various datasets. Our results  are structured into two parts: \textbf{(1)} We initially benchmark \methodname against other SSL methods using a popular self-supervised learning benchmark -- Lightly\cite{susmelj2020lightly} in Table \ref{tab:lightly-benchmark}. We follow the experimental protocol defined in the lightly benchmark \cite{susmelj2020lightly} after changing the backbone to a ViT. This comparison is pivotal to contextualize our method within the broader SSL landscape.  \textbf{(2)} Subsequently, we make an exhaustive list of comparisons and ablations with DINO, the SOTA ViT based SSL approach. For these experiments (Tables \ref{tab:dino-token-eval} to 7) we adapt the official DINO codebase, adhering to its hyper-parameter configurations for consistency. 

\noindent It is important to note that due to variations in experimental protocols and the number of training epochs, the results for identical datasets differ between Table \ref{tab:lightly-benchmark} and \ref{tab:maintab_sample}. Our SGC experiments are conducted using the same hyperparameters and setup as the baseline DINO experiments. This consistent setup ensures that any performance improvements can be directly attributed to the SGC method, rather than to differences in hyperparameters. For the implementation of graph layers, we use PyTorch Geometric \cite{Fey/Lenssen/2019}. In our SGC experiments, a two-layer graph neural network, followed by global mean pooling, is used to compute the feature representation for the SGC loss. Additionally, we configure the number of nearest neighbors for graph construction, $K$, to 20 and set the projection head dimension at 65536. Additional experimental details is reported in the Appendix.
\vspace{-0.3cm}

\subsection{Datasets}

Apart from the popular ImageNet-1k \cite{imagenet} dataset, our experimental framework encompasses a broad spectrum of datasets, from ImageNet subsets to domain-specific collections. This selection is intentional, our core objective diverges from introducing a novel SSL approach; instead, it focuses on \textit{\textbf{regularizing}} existing SSL methodologies that employ Vision Transformers (ViTs) as their backbone. To this end, our experiments span various datasets and settings, specifically tailored to contexts where the volume of data available for SSL pretraining varies. We test our method or a wide variety of datasets ranging from subsets of ImageNet to specialized collections specific to a particular problem with fine-grained sets of classes. Following this comprehensive evaluation, we also present a series of ablation studies, offering insights into the individual contributions of various components involved in our approach.



\noindent Apart form presenting results on the standard ImageNet-1K~\cite{imagenet} dataset, we employ subsets of the ImageNet dataset and several other datasets.
\textbf{ImageNet Subsets:} From the extensive ImageNet dataset with 1000 classes, we derived two smaller subsets: ImageNet-25 and ImageNet-100, containing 25 and 100 classes respectively. \textbf{RESISC-45} \cite{Cheng_2017}: This dataset is designed for Remote Sensing Image Scene Classification. It comprises 45 scene classes with 700 images each. 
\textbf{Food-101} \cite{bossard2014food}: This dataset features 101 food categories with each category having 750 training and 250 test images. 
\textbf{Caltech-256} \cite{griffin2007caltech}: Containing 257 diverse object categories, with each category containing atleast of 80 images, this datasets contains a wide range of objects, from grasshoppers to tuning forks.

\begin{table*}[h]
\centering
\begin{tabular}{@{}l|l|ccccccc<{\kern-\tabcolsep}}
\toprule
Dataset                       &  Method $\downarrow$ \% of Training data $\rightarrow$              & 1\%              & 5\%              & 10\%             & 25\%             & 50\%             & 75\%             & 100\%            \\ \midrule
\multirow{2}{*}{Food-101}    & {DINO \cite{caron2021emerging}}       & 36.89          & 52.89          & 58.30          & 62.52          & 65.34          & 66.69          & 67.46          \\
                             & \cellcolor{highlight} {DINO + {\methodname}} & \cellcolor{highlight} \textbf{42.25} & \cellcolor{highlight} \textbf{56.72} & \cellcolor{highlight} \textbf{61.43} & \cellcolor{highlight}\textbf{65.45} & \cellcolor{highlight}\textbf{67.92} & \cellcolor{highlight}\textbf{69.48} & \cellcolor{highlight}\textbf{70.23} \\ \midrule
\multirow{2}{*}{RESISC}      & {DINO \cite{caron2021emerging}}       & 56.97          & 74.92          & 78.43          & 73.76          & 83.43          & 84.22          & 84.73          \\
                             & \cellcolor{highlight}{DINO + {\methodname}} & \cellcolor{highlight}\textbf{61.94} & \cellcolor{highlight}\textbf{78.18} & \cellcolor{highlight}\textbf{81.18} & \cellcolor{highlight}\textbf{83.86} & \cellcolor{highlight}\cellcolor{highlight}\textbf{85.16} & \cellcolor{highlight}\textbf{86.03} & \cellcolor{highlight}\textbf{86.57} \\ \midrule
\multirow{2}{*}{Caltech-256} & {DINO \cite{caron2021emerging}}      & NA             & 22.03          & 28.82          & 36.61          & 40.89          & 42.99          & \textbf{47.41} \\
                             & \cellcolor{highlight}{DINO + {\methodname}} & \cellcolor{highlight}NA             &\cellcolor{highlight} \textbf{25.22} & \cellcolor{highlight}\textbf{30.63} & \cellcolor{highlight}\textbf{38.12} &\cellcolor{highlight} \textbf{43.39} &     \cellcolor{highlight}\textbf{45.58}            &\cellcolor{highlight} 47.36          \\ \bottomrule
\end{tabular}
\caption{\textit{\textbf{{\methodname} helps significantly even in low-data regimes.}}  Comparison of classification accuracies (\%) on various datasets using the baseline method (DINO - [CLS] token only) versus the proposed method (DINO + SGC) in low-data regimes}.
\label{tab:maintab_sample}
\end{table*}

\vspace{-0.3cm}
\subsection{Comparison with Existing SSL Approaches} 
This subsection delves into the performance of current state-of-the-art SSL methods when paired with a Vision Transformer (ViT) as the backbone. We employ the Lightly self-supervised learning benchmark~\cite{susmelj2020lightly} to conduct experiments on five prominent SSL methodologies:  BYOL~\cite{grill2020bootstrap}, SimCLR~\cite{chen2020simple}, Barlow Twins~\cite{zbontar2021barlow} and DINO~\cite{caron2021emerging}. Each method is trained across six datasets, with the outcomes detailed in Table \ref{tab:lightly-benchmark}.

\noindent Further, to assess the impact of our \methodname objective when combined with state-of-the-art SSL approach, we integrate \methodname with DINO framework~\cite{caron2021emerging}, owing to its widespread recognition and application in the field~\cite{meta-ai-forest}. Our approach, while designed for SSL methods utilizing a ViT backbone, boasts versatility, making it adaptable to any ViT-based SSL strategy. For each dataset, DINO, augmented with the \methodname objective, is pre-trained for 300 epochs. However, for ImageNet-1K, due to limitations in time and computational resources, both the baseline and our modified approach are pre-trained for only 100 epochs.

\noindent Table \ref{tab:lightly-benchmark} shows linear probing results on various datasets compared with existing ViT-based SSL methods. We observe that DINO + SGC outperforms DINO on all the datasets including the large-scale ImageNet-1k dataset. This is impressive given that DINO was highly tuned on ImageNet-1k dataset. We use the same hyperparameters as DINO to measure the true impact of SGC.




\subsection{Linear Probing in Limited Data Settings}
While linear probing is a standard evaluation protocol, its reliance on fully annotated datasets may not be entirely suitable for scenarios where labeled data is scarce. To address this, we further evaluate the quality of the representations learned by our method in limited data settings. In this setup, linear probing is performed using only a fraction of the training data, ranging from 1\% to 100\% of the available labeled samples. In datasets like Caltech-256, there are classes which contain only 80 samples, so for this this dataset results on 1\% are not applicable, so our table does not show them. Table \ref{tab:maintab_sample} showcases the results of this evaluation for three datasets: Food-101, RESISC, and Caltech-256. The results are reported for different fractions of the training data, and the performance of the baseline approach is compared with our {\methodname}-enhanced method on the test set.
\vspace{-0.3cm}

\noindent The results highlight the robustness of the representations learned by our method, especially in extremely limited data scenarios. For the Food-101 and RESISC datasets, our method consistently outperforms the baseline across all fractions of the training data. The proposed method achieves an impressive 5\% average gain in the setup with least amount of training data (only 1\% training data) on Food-101 and RESISC datasets.

\subsection{Ablation Study}

We conduct an extensive ablation study on the ImageNet-25 dataset to understand the role of each component in the framework.

\begin{table}[]
\resizebox{\linewidth}{!}{\begin{tabular}{>{\kern-\tabcolsep}l|cccc<{\kern-\tabcolsep}}
\toprule
                    & ImageNet-25 &  RESISC & Caltech-256 & Food-101 \\ \midrule
DINO                & 74.72       & 84.73  & \textbf{47.41}       & 67.46    \\
DINO + Patch Tokens & 73.44       &  80.01  & 35.52       &   64.89       \\
\rowcolor{highlight} DINO + SGC          & \textbf{78.64}       &  \textbf{86.57}  & 47.36      & \textbf{70.23}    \\ \bottomrule
\end{tabular}}
\caption{\textbf{\textit{Importance of graph construction and GNNs}}: We compare SGC to a baseline where we take the average of the patch tokens instead of pooled graph featuers and show that just using the average of patch tokens consistently degrades performance.} 
\label{tab:dino-token-eval}
\end{table}

\vspace{-0.5cm}
\paragraph{Importance of Graph Construction and Message passing using GNNs:}

To analyse the construction of the graph and message passing using GCNs, we compare it with a simple average of patch tokens without construction of the graph. We average the patch tokens and pass it to the projection head and compute the contrastive loss. The overall objective in this baseline setup (DINO + Patch) is similar to Eq. \ref{eq:overall-objective} but the second term is computed using the average of patch tokens instead of pooled graph features. Interestingly, we observe that directly using the average of patch tokens hurts DINO performance (Table \ref{tab:dino-token-eval}). This further reinforces the impact of constructing and capturing the relationship between the patches.


\vspace{-0.5cm}
\paragraph{K-Nearest Neighbor Graph Construction:}
A pivotal step in our approach is the construction of a K-Nearest Neighbor (KNN) graph using the patch tokens. The choice of $K$ plays a crucial role in determining the granularity and structure of the graph. In Table \ref{tab:k_value_ablation} , we experiment with different values of $K$ to ascertain the optimal number of neighbors that should be considered when building the graph. A high $K$ value ($K$=50) results in a dense graph thereby increasing the computation cost. We observe that as we increase the $K$ value, the performance gains start diminishing. We choose $K$ = 20 as the default value in order to balance between performance and computation cost.
\begin{table}[h]
\centering
\begin{tabular}{@{}l|ccccc@{}}
\toprule
$K$  & 3     & 5  & 10    & 20    & 30 \\ \midrule
Acc &  78.56 & 78.00 & 77.52 & 78.64 &   79.12    \\ \bottomrule
\end{tabular}
\caption{$K$-value (used for constructing the KNN-graph) ablation on ImageNet-25}
\label{tab:k_value_ablation}
\end{table}
\vspace{-0.7cm}
\paragraph{Graph Layers:}
Once the KNN graph is constructed, it is processed through a series of Graph Neural Network (GNN) layers. The choice of the GNN layer type can influence the feature transformation and messaging passing capabilities. We evaluate three prominent GNN layer types: GCN \cite{kipf2016semi},  Graph-SAGE \cite{hamilton2017inductive}, and GIN \cite{xu2018powerful}. This experiment sheds light on which GNN variant is best suited for our approach and how different layer types impact the learning of structural semantics. Results for this are shown in Table \ref{tab:gnn_layer_ablation}. While all the three GNN layers improve upon the baseline performance, SAGE and GCN performs much better than GIN layer.

\begin{table}[h]
\centering
\begin{tabular}{@{}c|cccc@{}}
\toprule
GCN Layer & No GCN & GCN &  SAGE  & GIN   \\ \midrule
Accuracy  & 74.72 & 78.64 & \textbf{78.72} & 77.68 \\ \bottomrule
\end{tabular}
\caption{Analysis of various GNN layers used in the \methodname module on ImageNet-25}
\label{tab:gnn_layer_ablation}
\end{table}
\vspace{-0.7cm}
\paragraph{Projection to High-Dimensional Space:}
Mirroring the projection technique used in DINO for the class token, we project the pooled graph features to a high-dimensional space. In Table \ref{tab:proj_dim_ablation}, we experiment with various projection dimensions, including 1024, 4096, 16384, 65536, and 262144. This ablation helps us understand the ideal dimensionality for the projection space, balancing computational efficiency and representational power. A large projection dimension leads to improved accuracy at the cost of higher computation.

\begin{table}[h]
\centering
\resizebox{\linewidth}{!}{
\begin{tabular}{@{}l|cccccc@{}}
\toprule
Dimension & 512 & 1024  & 4096  & 16384 & 65536 & 262144 \\ \midrule
Accuracy  & 75.28 &  75.68 & 75.84 & 76.96 & 78.64 & 79.04  \\ \bottomrule
\end{tabular}
}
\caption{Ablation of projection head dimension on ImageNet-25}
\label{tab:proj_dim_ablation}

\end{table}

\vspace{-0.5cm}
\paragraph{Weighted Combination of Losses:}
Our overall objective function, as depicted in Eq. \ref{eq:overall-objective}, is a combination of two distinct losses. To understand the role of each loss, we assign a weight $\alpha$ that governs weight associated with the consistency between the projected features of the class token for both augmented views. As shown in Eq \ref{eq:overall-objective}, $\beta$ determines the weightage of the {\methodname} consistency. To dissect the contributions of each loss, we experiment with various configurations:

\begin{compactitem}
\item 
$\alpha=1$ and $\beta=0$  This configuration essentially represents the baseline method, relying solely on the consistency of the class token.
\item 
$\alpha=0$ and $\beta=1$: Here, we evaluate the performance when only the {\methodname} consistency is employed, excluding the class token consistency.
\item Varying $\beta$ values, keeping $\alpha$ fixed at 1, we experiment with different $\beta$ values, including 0, 0.1, 0.3, 0.5, and 1. This helps us understand the influence of the {\methodname} consistency when used in conjunction with the class token consistency
\end{compactitem}

\begin{table}[h]
\centering
\resizebox{\linewidth}{!}{\begin{tabular}{@{}c|cccccc@{}}
\toprule
$\alpha$ ({[}CLS{]}) & 1     & 0     & 1     & 1     & 1     & 1     \\ \midrule
$\beta$ ({\methodname})        & 0     & 1     & 0.1   & 0.3   & 0.5   & 1     \\ \midrule
Accuracy          & 74.64 & 69.04 & 77.76 & \textbf{79.28} & 78.48 & 78.64 \\ \bottomrule
\end{tabular}}
\caption{Ablation of the $\alpha$ (weight for the class token consistency loss) and $\beta$ values in the overall objective function on  ImageNet-25}
\label{tab:alpha_beta_ablation}
\end{table}

\vspace{-0.5cm}
\noindent Our findings in Table \ref{tab:alpha_beta_ablation} indicate that a configuration of $\alpha=1$ and $\beta=0.3$ yields the best results, reinforcing the notion that {\methodname} is most effective as a regularizer rather than a standalone objective.

\section{Conclusion and Future Work}

In this work, we focused on the problem of regularizing Vision Transformer (ViT)-based Self-Supervised Learning (SSL) techniques. We introduced a novel method, termed \methodnamefull (\methodname), which ingeniously reconceptualizes images as graphs to infuse relational inductive biases into the SSL framework. This approach leverages the underexploited patch tokens of ViTs, capturing intricate relationships between image patches akin to the human visual perception system. We demonstrated that our {\methodname} loss acts as a powerful regularizer, significantly enhancing the quality of learned representations on a broad range of datasets. Empirical results, including extensive ablation studies, have confirmed that our approach outperforms existing methods by 5-10\% under limited data scenarios, which is an impactful contribution to the SSL community. We hope that this work serves as a compelling demonstration of the potential of graph-based approaches in SSL, stimulating further research into the architectural, algorithmic, and theoretical aspects of this promising intersection between graph theory and self-supervised learning.

{
    \small
    \bibliographystyle{ieeenat_fullname}
    \bibliography{references}
}

\newpage 
\section*{\centering }
\newpage 
\section*{\centering Appendix}
\noindent This appendix provides supplementary details that were omitted from the main paper due to space constraints. It focuses on a more in-depth description of our experimental setup, supplemented by additional results. The key aspects covered in this appendix are summarized as follows:

\begin{compactitem}
    \item \textbf{Experimental Details:} We outline the specifics of the baseline experiments we performed with respect to DINO~\cite{caron2021emerging} in Tables 2 to 7 of the main paper, our proposed method (\methodname), and the linear evaluation protocol. This includes hyperparameters, training strategies, and other relevant details
        
    \item \textbf{Dataset Overview:} We present a summary of the datasets used in our experiments, specifying the number of classes and images in the training and test sets for each dataset.
        
    \item \textbf{Alternative Graph Construction:} We explore the use of cosine similarity as an alternative to Euclidean distance for KNN graph construction in our method.

    \item \textbf{Effectiveness of Pre-training:} We discuss the comparative effectiveness of pre-training on small-scale datasets followed by linear probing, versus training a model from scratch, especially in limited-data scenarios.

\end{compactitem}

\noindent This appendix aims to complement the main paper by offering a comprehensive understanding of the experimental setup and results that support our findings.

\subsection{Experimental Details}
\noindent Our experiments are conducted on NVIDIA A30 GPUs. Our codebase is built on PyTorch \cite{paszke2019pytorch}. For Tables 2 to 7 of the main paper, we use the official DINO implementation \cite{caron2021emerging}\footnote{https://github.com/facebookresearch/dino} as a baseline and implement our \methodnamefull method atop this baseline. Upon acceptance, we will release our code and checkpoints for public use.
\\
\\
\textbf{DINO Baseline:}
We replicate the experiments using the hyperparameters specified in the original DINO paper. Our setup involves a batch size of 128 and a learning rate of \(5 \times 10^{-4}\). We employ the AdamW optimizer \cite{loshchilov2017decoupled} for training the model, with a cosine annealing scheduler and a 10-epoch warmup period. For the Vision Transformer (ViT) backbone, we utilize a patch size of 16. The Exponential Moving Average (EMA) parameter for the teacher network is set to 0.996.
\\
\\
\textbf{\methodnamefull (\methodname) Experiments:}\\
Our SGC experiments are conducted using the same hyperparameters and setup as the baseline DINO experiments. This consistent setup ensures that any performance improvements can be directly attributed to the SGC method, rather than to differences in hyperparameters. For the implementation of graph layers, we use PyTorch Geometric \cite{Fey/Lenssen/2019}. In our SGC experiments, a two-layer graph neural network, followed by global mean pooling, is used to compute the feature representation for the SGC loss. Mathematically, the SGC loss (as defined in Eq \ref{eq:sgc-eq}) is designed to be equivalent to the DINO loss but leverages the pooled graph features instead.
\\
\\
\textbf{Linear Evaluation Protocol:}
In alignment with the standard protocol for self-supervised learning (SSL), we train a linear classifier on top of the frozen pre-trained representations. For input to the linear classifier, we concatenate the class token of the last four blocks. The training employs the SGD optimizer with a momentum of 0.9, no weight decay, and a learning rate of \(5 \times 10^{-4}\). The linear classifier is trained for a total of 100 epochs with a batch size of 128. Additionally, we apply a Random Resized Crop augmentation during the training phase, adhering to common practices in linear evaluation protocols \cite{loshchilov2017decoupled}.

\subsection{Datasets Used for Evaluation}
In this section, we provide a comprehensive summary of the datasets employed to evaluate the effectiveness of our proposed method. Table~\ref{tab:abl_data_100_summary} outlines these datasets, specifying the number of classes and the number of images designated for training and testing in each dataset. Notably, these datasets correspond to those referenced in Table~\ref{tab:lightly-benchmark} of the main paper.
\begin{table}[h]
\resizebox{\linewidth}{!}{\begin{tabular}{@{}l|c|c|c@{}}
\toprule
Dataset      & \# classes & \# Images (training) & \# Images (test) \\ \midrule
ImageNet-25  & 25         & 32285                & 1250             \\
ImageNet-100 & 100        & 130000               & 5000             \\
ImageNet-1K & 1000        & 1281167               & 50000             \\
Caltech      & 257        & 24385                & 6222             \\
RESISC       & 45         & 25200                & 6300             \\
Food         & 101        & 75750                & 25250            \\ \bottomrule
\end{tabular}}
\caption{Summary of datasets used for the evaluation of our method, indicating the number of classes, and the number of images in the training and test sets, as referenced in Table \ref{tab:lightly-benchmark} of the main paper.}
\label{tab:abl_data_100_summary}
\end{table}
\noindent The specific classes selected from the ImageNet-25 and ImageNet-100 datasets for our experiments will be listed in the accompanying the codebase (which we will release upon acceptance)

\noindent In Table \ref{tab:maintab_sample} of the main paper, we evaluate the effectiveness of our method under limited-sample linear evaluation scenarios. Specifically, we train the model using subsets of the original training set, sized at 1\%, 5\%, 10\%, 25\%, 50\%, 75\%, and 100\%. Testing is performed using the original test set. Table \ref{tab:abl_data_sample_summary} provides a summary of the number of training images for each training split along with the number of test images used. 

\begin{table}[h]
\resizebox{\linewidth}{!}{\begin{tabular}{l|c|ccccccc|c}
\hline
Dataset & \# classes & \multicolumn{7}{c|}{\# Images (training)}             & \# Images (test) \\ \toprule
        &            & 1\%     & 5\%   & 10\%  & 25\%   & 50\%   & 75\%   & 100\%  &                  \\ \midrule
Caltech & 257        & NA (60) & 1103  & 2317  & 6027   & 12144  & 18214  & 24385  & 6222             \\
RESISC  & 45         & 225     & 1240  & 2502  & 6285   & 12591  & 18879  & 25200  & 6300             \\
Food    & 101        & 707     & 3737  & 7575  & 18887  & 37875  & 56762  & 75750  & 25250            \\ \bottomrule
\end{tabular}}
\caption{Summary of datasets used for limited-sample linear evaluation of our method. The table indicates the number of classes, the number of images in various training set sizes (expressed as percentages of the original training set), and the number of images in the test sets, as referenced in Table \ref{tab:maintab_sample} of the main paper.}
\label{tab:abl_data_sample_summary}
\end{table}

\subsection{Alternative Graph Construction}
\paragraph{Cosine Similarity for Graph Construction:}
In the experiments presented in the main paper, we construct a \( k \)-nearest neighbors (KNN) graph using the patch tokens of the images. The Euclidean distance metric is employed for this purpose. In contrast, Table~\ref{tab:cosine_k_value_ablation} presents results from an alternative experiment conducted on the ImageNet-25 dataset, where we construct the KNN graph using cosine similarity as the distance metric instead of Euclidean distance. We observe that, with cosine similarity as the distance metric, setting \( K = 3 \) yields the best performance. Increasing the value of \( K \) beyond this point often leads to a decrease in performance.

\begin{table}[h]
\centering
\begin{tabular}{@{}l|cccc@{}}
\toprule
\( K \)  & 3     & 5  & 10    & 20     \\ \midrule
Accuracy (\%) & 79.20 & 78.72 & 78.48 & 77.26     \\ \bottomrule
\end{tabular}
\caption{Performance ablation on ImageNet-25 with different values of \( K \) when using cosine similarity for KNN graph construction.}
\label{tab:cosine_k_value_ablation}
\end{table}

\begin{table}
\resizebox{\linewidth}{!}{\begin{tabular}{@{}l|ccccccc@{}}
\toprule
Dataset $\downarrow$ $/$  \% of Training data $\rightarrow$ & 1    & 5     & 10    & 25    & 50    & 75    & 100   \\ \midrule
Food-101                        & 1.48 & 4.9   & 6.76  & 13.87 & 23.73 & 31.74 & 38.12 \\
RESISC                          & 6.56 & 15.42 & 17.75 & 29.96 & 34.03 & 52.35 & 59.07 \\
Caltech-256                     & NA   & 6.33  & 7.53  & 11.01 & 17.51 & 22.27 & 27.79 \\ \bottomrule
\end{tabular}}
\caption{Performance of a Vision Transformer (ViT) model trained from scratch under various data regimes, expressed as percentages of the original training set size.}
\label{tab:abl_supervised_scratch}
\end{table}

\subsection{Effectiveness of Pre-training}

\paragraph{Training from Scratch on Small-Scale Datasets:}
In the main paper, all experiments conducted on small-scale datasets (any dataset other than ImageNet-1K) involve pre-training on unlabelled samples, followed by linear evaluation. A valid question that arises is: why not train directly on the small-scale dataset itself? To empirically demonstrate the efficacy of pre-training on small-scale datasets followed by linear probing, as compared to training from scratch, we conduct a set of experiments. In these experiments, we train the Vision Transformer (ViT) model directly on small-scale datasets, without leveraging any unlabelled data. Table~\ref{tab:abl_supervised_scratch} presents the results of training a ViT from scratch under limited data conditions. Comparing the results in Table \ref{tab:maintab_sample} with those in Table~\ref{tab:abl_supervised_scratch}, it becomes evident that pre-training proves beneficial not only in low-data scenarios, but also when the entire training dataset is utilized.


\end{document}